%% file: bare_jrnl.tex
\documentclass[journal]{IEEEtran}

\usepackage{microtype}

\usepackage{graphicx}
\usepackage{amsmath}
\usepackage{amsfonts}
\usepackage{mathtools}
\usepackage{diagbox}
\usepackage{multirow}
\usepackage{float}
\usepackage{icomma}
\usepackage[range-phrase=--]{siunitx}
\usepackage{algorithm, algpseudocode}
\usepackage{parskip}
\usepackage{tabularx}
\usepackage{booktabs}

\usepackage{xargs}    
\usepackage[pdftex,dvipsnames]{xcolor} 
\usepackage[colorinlistoftodos,prependcaption,textsize=tiny]{todonotes}

\newcommandx{\improvement}[2][1=]{\todo[linecolor=Plum,backgroundcolor=Plum!25,bordercolor=Plum,#1]{#2}}

\usepackage{glossaries}
\newacronym{MLC}{MLC}{multi-label scene classification}
\newacronym{SLC}{SLC}{single-label classification}
\newacronym{RS}{RS}{remote sensing}
\newacronym{CV}{CV}{computer vision}
\newacronym{LP}{LP}{label propagation}
\newacronym{xAI}{xAI}{explainable artificial intelligence}
\newacronym{GAN}{GAN}{generative adversarial network}
\newacronym{SAR}{SAR}{synthetic aperture radar}
\newacronym{CAMs}{CAMs}{class activation maps}
\newacronym{LULC}{LULC}{land use land cover}
\newacronym{RRC}{RRC}{RandomResizeCrop}
\newacronym{RR}{RR}{RandomRotate}
\newacronym{ELR}{ELR}{early-learning regularization}

\makeatletter
\newcount\SOUL@minus 
\makeatother

\usepackage{color, soul}

\usepackage{etoolbox}
\usepackage{overpic}

\renewcommand{\arraystretch}{1.4}%

\def\subcaptionabove{subcaptionabove}
\def\subcaptionbelow{subcaptionbelow}
\def\subcaptionstyle{subcaptionbelow}

\usepackage{cite}
\usepackage{hyperref}
\usepackage[
    singlelinecheck=false, 
    font=small,
]{caption}
\usepackage[slc=off, subrefformat=parens]{subcaption}

\ifx\subcaptionstyle\subcaptionbelow
\else
\fi

\newcommand{\dynamiccaption}[2]{
    \ifx\subcaptionstyle\subcaptionabove
        \setlength{\belowcaptionskip}{10pt}
        #1
    \fi
    #2
    \ifx\subcaptionstyle\subcaptionbelow
        #1
    \fi
}

\usepackage[capitalise,noabbrev]{cleveref}

\makeatletter
\patchcmd{\@makecaption}
  {\scshape}
  {}
  {}
  {}
\makeatletter
\patchcmd{\@makecaption}
  {\\}
  {.\ }
  {}
  {}
\makeatother

\begin{document}

\title{\title{Noise-Adaptive Regularization for Robust Multi-Label Remote Sensing Image Classification}}

\author{Tom~Burgert,~\IEEEmembership{~Member,~IEEE,}
        Julia Henkel,~\IEEEmembership{~Member,~IEEE,}
        and~Begüm~Demir,~\IEEEmembership{Senior~Member,~IEEE}
\thanks{T. Burgert, J. Henkel, and B. Demir are with the Berlin Institute for the Foundations of Learning and Data (BIFOLD) and with the Faculty of Electrical Engineering and Computer Science, Technische Universit{\"a}t Berlin, 10623 Berlin,
Germany (emails: t.burgert@tu-berlin.de, henkel@campus.tu-berlin.de,  demir@tu-berlin.de).}
}

\markboth{Journal of \LaTeX\ Class Files,~Vol.~14, No.~8, August~2015}%
{Shell \MakeLowercase{\textit{et al.}}: Bare Demo of IEEEtran.cls for IEEE Journals}

\maketitle


\begin{abstract}
The development of reliable methods for multi-label classification (MLC) has become a prominent research direction in remote sensing (RS). As the scale of RS data continues to expand, annotation procedures increasingly rely on thematic products or crowdsourced procedures to reduce the cost of manual annotation. While cost-effective, these strategies often introduce multi-label noise in the form of partially incorrect annotations. In MLC, label noise arises as additive noise, subtractive noise, or a combination of both in the form of mixed noise. Previous work has largely overlooked this distinction and commonly treats noisy annotations as supervised signals, lacking mechanisms that explicitly adapt learning behavior to different noise types. To address this limitation, we propose NAR, a noise-adaptive regularization method that explicitly distinguishes between additive and subtractive noise within a semi-supervised learning framework. NAR employs a confidence-based label handling mechanism that dynamically retains label entries with high confidence, temporarily deactivates entries with moderate confidence, and corrects low confidence entries via flipping. This selective attenuation of supervision is integrated with early-learning regularization (ELR) to stabilize training and mitigate overfitting to corrupted labels. Experiments across additive, subtractive, and mixed noise scenarios demonstrate that NAR consistently improves robustness compared with existing methods. Performance improvements are most pronounced under subtractive and mixed noise, indicating that adaptive suppression and selective correction of noisy supervision provide an effective strategy for noise robust learning in RS MLC.
\end{abstract}


\IEEEpeerreviewmaketitle

\input{sections/1-introduction}
\input{sections/2-related}
\input{sections/3-methods}
\input{sections/4-dataset_design}
\input{sections/5-experimental_results}
\input{sections/6-conclusion}

\ifCLASSOPTIONcaptionsoff
  \newpage
\fi

\bibliography{references}
\bibliographystyle{ieeetr}
\vfill

\end{document}

%% file: sections/1-introduction.tex
\section{Introduction}\label{sec:introduction} \IEEEPARstart{R}{ecent} advances in satellite technology have led to a substantial growth of \gls{RS} image archives, facilitating detailed analysis of Earth observation data at scale. In particular, \gls{MLC} has emerged as a suitable formulation for analyzing \gls{RS} images, as they typically cover geographical areas that often contain multiple land cover classes within a single scene. In the past few years, \gls{MLC} works have adopted deep learning approaches that significantly improve classification performance \cite{sumbul_deep_2020}, \cite{sobti_ensemv3x_2021}, \cite{li_multi-label_2020}. However, supervised deep learning typically requires large amounts of accurately annotated training data, which makes the manual labeling process both expensive and time-consuming. To address this limitation, alternative annotation strategies, including crowdsourcing and the use of thematic products such as Corine Land Cover \cite{buttner_corine_2004}, GLC2000 \cite{bartholome_glc_2002}, and GlobCover \cite{arino_global_2012}, are increasingly employed in \gls{RS} \cite{paris_novel_2020}, \cite{sumbul_bigearthnet_2019}, \cite{clasen_reben_2025}. Although cost-effective, these annotation strategies often introduce multi-label noise (i.e., partially incorrect labels for each image). Crowdsourced labels are prone to human error and inconsistency, whereas thematic products may include outdated or spatially inaccurate mappings.

In \gls{MLC}, label noise arises primarily as additive noise or subtractive noise, and often as a combination of both in the form of mixed noise \cite{burgert_effects_2022}. Additive noise occurs when absent classes are incorrectly annotated as present, whereas subtractive noise occurs when present classes are omitted from annotations. Previous \gls{RS} works have often failed to distinguish between these types, commonly applying uniform noise injection schemes for evaluating robustness that disproportionately emphasize additive noise and underrepresents the influence of subtractive noise \cite{hua_learning_2020}. Burgert et al.\cite{burgert_effects_2022} address this limitation by proposing a noise injection strategy that independently controls both components, enabling systematic analysis of additive and subtractive noise in isolation as well as their joint effect under mixed noise. Their results show that these noise types affect model robustness in qualitatively different ways, with subtractive noise having a particularly detrimental effect.

Although the formal distinction between additive and subtractive noise has established a clearer analytical basis to study robustness under label noise in \gls{MLC}, this conceptual progress has not led to corresponding methodological advances. Existing approaches in \gls{MLC} \cite{hua_learning_2020}, \cite{aksoy_multi-label_2022}, \cite{ridnik_asymmetric_2021}, \cite{song_toward_2024}
typically treat noisy annotations as supervised signals and lack mechanisms that explicitly adapt learning behavior to different noise types. This is restrictive, since noisy multi-label annotations may combine informative and misleading supervision within the same sample, making uniform treatment of all label entries inadequate. Semi-supervised learning provides a principled framework for addressing this ambiguity, as it explicitly permits selective attenuation of supervision by reinterpreting unreliable label entries as unlabeled, while still exploiting their underlying data structure. In \gls{SLC}, semi-supervised strategies have been proven highly effective for noise-robust training through mechanisms such as confidence-based sample selection \cite{jiang_mentornet_2018}, \cite{han_co-teaching_2018}, consistency regularization \cite{nguyen_self_2020}, \cite{li_dividemix_2019}, and pseudo-label refinement \cite{huang_self-adaptive_2020}, \cite{liu_early-learning_2020}. Nevertheless, extending these strategies to the multi-label setting is non-trivial, as correlated label structures, class imbalances, and the coexistence of multiple noise types each influence model behavior in distinct ways \cite{burgert_effects_2022}. In particular, \gls{RS} \gls{MLC} commonly suffers from two forms of imbalance. First, some land cover classes occur substantially more frequently than others, leading to class imbalance. Second, because each class is absent in the majority of images, positive label entries constitute only a small fraction of the overall label matrix, resulting in a strong present--absent imbalance. These properties complicate the identification of noisy label entries and increase the risk that incorrect supervision dominates the learning process. To address the above mentioned challenges, we introduce NAR, a novel noise-adaptive regularization method that integrates confidence-based label correction with a multi-label adaptation of \gls{ELR}. Our method distinguishes between additive and subtractive noise by dynamically adapting label handling according to model confidence. Label entries with high confidence are retained, label entries with moderate confidence are temporarily deactivated, while those with low confidence are corrected via flipping. This enables a semi-supervised, noise-type–adaptive regularization process that selectively suppresses unreliable supervision while preserving informative samples.

In summary, our contributions are the following:

\begin{itemize} 
\item We provide a systematic analysis showing that confidence-based label flipping strategies developed for \gls{SLC} do not directly extend to the multi-label setting, due to asymmetric effects of additive and subtractive noise.
\item We propose NAR, the first noise-type–adaptive semi-supervised regularization method for \gls{MLC} that combines a multi-label adaptation of \gls{ELR} with a dynamic confidence-based label handling mechanism. In detail, the method employs a three-state mechanism that assesses the reliability of each label entry, retaining entries with high confidence, deactivating those with moderate confidence, and correcting (i.e., flipping) those with low confidence. 
\item We demonstrate that NAR consistently improves robustness across three multi-label \gls{RS} datasets and three noise scenarios (additive, subtractive, mixed), outperforming existing state-of-the-art methods across various noise rates.
\end{itemize}

The remainder of this paper is structured as follows: Section \ref{related} reviews related work on label noise robustness in \gls{SLC} as a source of established noise robust learning strategies, followed by existing approaches in \gls{MLC}. Section \ref{methods} presents the proposed NAR, which integrates confidence-based label correction and \gls{ELR} within a semi-supervised training framework. Section \ref{dataset_setup} describes the considered multi-label datasets and the experimental setup, while the experimental results are presented in Section \ref{experimental_results}. Finally, in Section \ref{conclusion}, the conclusion of the work is drawn.


%% file: sections/2-related.tex
\section{Related Work}
\label{related}
Research on learning under label noise has developed along multiple methodological directions in recent years, with substantial progress in \gls{SLC}, primarily within the \gls{CV} community, and growing interest in \gls{MLC} across both \gls{RS} and \gls{CV}. In \gls{SLC}, extensive theoretical and methodological foundations for noise robust learning have been established, including the systematic integration of semi-supervised principles. In contrast, \gls{MLC} introduces fundamentally different noise characteristics due to the coexistence of multiple class annotations per image, which complicates both theoretical analysis and empirical evaluation. 
As a result, noise robust learning in \gls{MLC} remains an evolving research area, with methodological choices that have not yet converged to a unified set of practices. The following review summarizes representative methods across these communities, with \Cref{sec:rw_slc} focusing on noise-robust learning in \gls{SLC} \gls{CV}, and \Cref{sec:rw_mlc} examining how these advances extend to \gls{MLC} in both \gls{RS} and \gls{CV}.

\subsection{Single-Label Noise Robust Methods}\label{sec:rw_slc}

Early research on learning with noisy labels often has assumed the existence of a small clean subset that could guide the training process. Hendrycks et al.\cite{hendrycks_using_2018} use clean data to estimate a noise transition matrix for label correction, while Lee et al. \cite{lee_cleannet_2018} construct class prototypes in feature space to weight noisy samples by similarity to clean ones. In parallel, other works pursued noise robustness directly through model and loss design without relying on clean supervision. Prominent examples include noise-aware architectures \cite{song_collaborative_2018}, \cite{sukhbaatar_training_2015} and noise robust loss formulations \cite{ghosh_robust_2017}, \cite{zhang_generalized_2018}.

Building on these early efforts, subsequent methods increasingly incorporate principles from semi-supervised learning, operating under the assumption that a subset of samples is likely clean and can guide the learning process. These methods typically identify such samples through prediction confidence or training dynamics, followed by mechanisms such as reweighting, sample filtering, consistency regularization, or co-training. Curriculum-based methods structure the learning order to limit the impact of label noise. CurriculumNet \cite{guo_curriculumnet_2018} ranks samples by feature-space complexity and progressively trains from simple to complex examples, while MentorNet \cite{jiang_mentornet_2018} learns a data-driven curriculum that adaptively weights samples by estimated reliability. Collaborative and co-teaching strategies employ multiple networks to mitigate label noise effects. Decoupling \cite{malach_decoupling_2017} trains two networks that update only on samples where their predictions disagree, preventing mutual reinforcement of incorrect labels. Co-Teaching \cite{han_co-teaching_2018} extends this idea by exchanging small-loss samples between networks, and JoCoR \cite{wei_combating_2020} further couples them through a co-regularization term that enforces agreement while maintaining small-loss selection. Finally, SELF \cite{nguyen_self_2020} integrates filtering and consistency regularization within a single-network framework by forming a temporal self-ensemble that averages predictions across epochs to identify inconsistent labels. Samples identified as noisy are excluded from the supervised loss but retained under a consistency regularization term applied to perturbed inputs.

While these methods implicitly adopt semi-supervised principles by relying on likely clean subsets to guide training, more recent research explicitly formulates learning with noisy labels as a semi-supervised problem. In these approaches, confident samples serve as labeled data, and uncertain or mislabeled ones are treated as unlabeled data that contribute through consistency or pseudo-labeling mechanisms. Han et al. \cite{han_deep_2019} propose a deep self-learning framework that iteratively refines labels via feature-space prototypes and trains the model in an end-to-end self-labeling loop. Self-adaptive training (SAT) \cite{huang_self-adaptive_2020} updates labels dynamically using the exponential moving average of model predictions, thereby enforcing temporal consistency without additional supervision. ELR \cite{liu_early-learning_2020} exploits the early-learning phase of neural networks to derive semi-supervised target probabilities and regularize the model to prevent memorization of corrupted labels. DivideMix \cite{li_learning_2019} partitions the data into labeled and unlabeled subsets through a mixture model on the loss distribution and jointly trains two networks using a semi-supervised MixMatch \cite{berthelot_mixmatch_2019} objective. Extending this work, Zheltonozhskii et al.\cite{zheltonozhskii_contrast_2022} propose contrast to divide (C2D), which incorporates a self-supervised contrastive pretraining stage before semi-supervised noise robust training, showing that self-supervised pretrained representations mitigate warm-up memorization and improve separation between clean and noisy samples under severe corruption.

\subsection{Multi-Label Noise Robust Methods}\label{sec:rw_mlc}

Although research on noise-robust learning is well established in \gls{SLC}, with standardized benchmarks and clear definitions of label noise, the work in \gls{MLC} remains heterogeneous and conceptually fragmented. According to the definition introduced by Burgert et al. \cite{burgert_effects_2022}, label noise in \gls{MLC} can manifest as subtractive noise, where present classes are unannotated, or additive noise, where absent classes are incorrectly annotated as present. Several related research areas have implicitly addressed these forms without explicitly simulating noise to evaluate robustness to label corruption. In multi-label learning with missing labels, the learning problem is defined around incomplete annotations, where missing class assignments correspond to subtractive noise and are typically treated as negatives \cite{durand_learning_2019}. Early works such as Bucak et al.\cite{bucak_multi-label_2011} and Jain et al. \cite{jain_extreme_2016} approach this problem through ranking-based formulations, while Durand et al. \cite{durand_learning_2019} adapted the binary cross-entropy loss to the proportion of known labels and incorporated label correlations through a graph neural network. Conversely, partial multi-label learning is formulated around candidate label sets that include extraneous classes, effectively introducing additive noise through spurious positive annotations. Xie et al.\cite{xie_partial_2021} mitigate this issue by ranking candidate labels using confidence-weighted losses informed by label correlations and feature prototypes.

Only a limited number of studies have explicitly examined label noise in \gls{MLC}. Early works such as Veit et al. \cite{veit_learning_2017}, Hu et al. \cite{hu_multi-label_2018}, and Inoue et al. \cite{inoue_multi-label_2017} introduce auxiliary label-cleaning networks trained on small clean subsets to supervise classifiers trained on noisy data. These approaches depend on the availability of clean annotations and are evaluated only on datasets affected by inherent and uncontrolled label noise, without systematic control over the corruption process. Zhao and Gomes \cite{zhao_evaluating_2021} later provide the first controlled study of additive and subtractive noise by introducing an encoder–decoder architecture with an attention-based graph neural network and uniform noise injection, while Kumar et al. \cite{kumar_robust_2020} offered theoretical evidence that the mean absolute error is robust to symmetric corruption, though this assumption rarely holds in sparse multi-label settings. More recent methods have adapted noise-robust principles to \gls{MLC} by modifying loss formulations and training dynamics. The asymmetric loss (ASL) \cite{ridnik_asymmetric_2021} modulates gradient contributions from positive and negative labels to counteract class imbalance and reduce the impact of mislabeled negatives, whereas BalanceMix \cite{song_toward_2024} employs mixup-inspired data augmentation to oversample minority labels and refines label assignments at the label-wise level, categorizing them as clean, re-labeled, or ambiguous.

In \gls{RS}, prior studies have approached label noise in \gls{MLC} without a rigorous definition or principled evaluation of individual noise effects. Hua et al. \cite{hua_learning_2020} regularize network predictions using a semantic label correlation matrix derived from word embeddings while assuming uniform corruption across labels. Due to the inherent class imbalance of multi-label datasets, this assumption results predominantly in additive noise, with very few corrupted positive labels. Aksoy et al. \cite{aksoy_multi-label_2022} introduce a collaborative dual network framework that combines a discrepancy and a consistency loss. They simulate noise only at the sample level rather than per label, which produces a mixture of entirely clean and mostly corrupted samples (i.e., multiple noisy label entries per sample) that favors their method under evaluation. Kondylatos et al. \cite{kondylatos_probabilistic_2025} apply probabilistic deep learning to \gls{RS} images, estimating input-dependent uncertainty to handle inherently noisy datasets, yet without explicitly modeling noise or analyzing distinct noise types. Burgert et al. \cite{burgert_effects_2022} address these conceptual and methodological limitations by first formalizing a taxonomy of additive, subtractive, and mixed multi-label noise, and then adapting established single-label noise-robust algorithms, including SAT \cite{huang_self-adaptive_2020}, ELR \cite{liu_early-learning_2020}, and JoCoR \cite{wei_combating_2020}, to \gls{MLC}.

%% file: sections/3-methods.tex
\section{Methodology}\label{methods}

\begin{figure*}[t]
    \centering
    \includegraphics[width=0.9\linewidth]{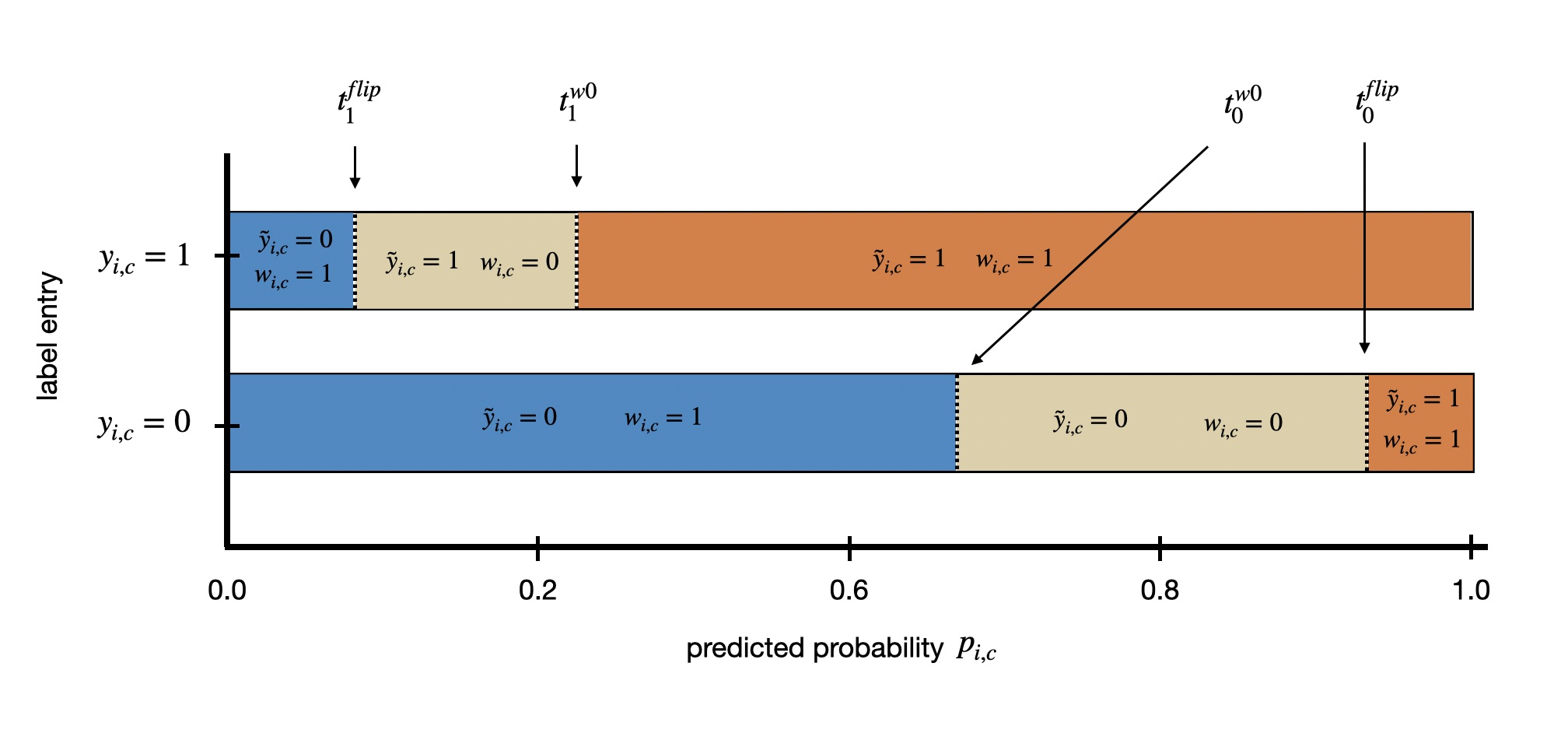}
    \caption{Schematic illustration of the confidence-based label handling mechanism. Each label entry (i.e., 0 or 1) is handled via one of three states: (1) label entry retaining of likely clean entries, when $y_{i,c}$ is close to $p_{i,c}$, (2) label entry deactivation via setting weight $w_{i,c}$ to 0, when $p_{i,c}$ is neither close to 0 or 1, (3) label entry flipping when $y_{i,c}$ is close to $1 - p_{i,c}$. Resulting corrected label entries $\tilde{y}_{i,c} = 0$ if colored in blue, $\tilde{y}_{i,c} = 1$ if colored in orange, and deactivated when colored in beige.}
    \label{fig:method_overview}
\end{figure*}

In this section, we introduce NAR, a novel noise-adaptive regularization method for \gls{MLC} under noisy supervision. To the best of our knowledge, NAR is the first approach that explicitly distinguishes between additive and subtractive label noise at the label entry level using semi-supervised learning strategies. NAR improves robustness to label corruption by adapting the handling of individual label entries according to their estimated noise type and reliability. Specifically, NAR treats noisy supervision as partially reliable, where each label entry may be retained, temporarily deactivated, or corrected depending on model confidence. Label entries with high confidence of correctness are retained, entries with intermediate uncertainty are deactivated and treated as unlabeled, and entries with high confidence of corruption are flipped and used as pseudo labels. This three-state formulation transforms noisy supervised learning into a semi-supervised process that dynamically adapts both supervision strength and direction at the label entry level. NAR integrates this confidence-based mechanism with \gls{ELR} adapted to \gls{MLC} to stabilize training and suppress overfitting to corrupted labels.

Let the training set be defined as
\[
\mathcal{D} = \{(\mathbf{x}_1, \mathbf{y}_1), \dots, (\mathbf{x}_N, \mathbf{y}_N)\},
\]
where each image \(\mathbf{x}_i\) is associated with a multi-label vector \(\mathbf{y}_i \in \{0, 1\}^C\), with each label entry $c$ indicating the presence or absence of each class \(c \in \{1, \dots, C\}\). We follow Burgert et al. \cite{burgert_effects_2022} and define additive noise as false positives (\(y_{i,c} = 1\) when class \(c\) is absent) and subtractive noise as false negatives (\(y_{i,c} = 0\) when class \(c\) is present).

Let \(f(\cdot; \theta)\) be a neural network with parameters \(\theta\), predicting per-class probabilities (i.e., confidences) \(\mathbf{p}_i = f(\mathbf{x}_i; \theta)\), where each \(p_{i,c} \in [0, 1]\) denotes the predicted probability for class \(c\). The basic learning objective is the binary cross-entropy (BCE) loss:
\begin{equation}
L_{\text{BCE}}(\theta) = \frac{1}{nC} \sum_{i=1}^n \sum_{c=1}^C \left[ y_{i,c} \log(p_{i,c}) + (1 - y_{i,c}) \log(1 - p_{i,c}) \right]
\end{equation}

The following subsections describe the adaptation of \gls{ELR} to the multi-label setting and the proposed confidence-based label handling mechanism that enables noise-type–adaptive semi-supervised learning.

\subsection{Early-learning Regularization}

Neural networks tend to fit clean labels early in training before memorizing noise. To exploit this property, we adapt the \gls{ELR} loss \cite{liu_early-learning_2020} to \gls{MLC}. ELR penalizes deviation from early predictions, thus stabilizing training and suppressing the influence of corrupted labels. In the original single-label formulation, the regularization term is as follows:
\begin{equation}
L_{\text{ELR}}(\theta) = \frac{\lambda}{N} \sum_{i=1}^N \log \left( 1 - \langle \mathbf{p}_i, \mathbf{y}_i \rangle \right),
\end{equation}

where $\lambda$ controls the regularization strength. 

We extend this formulation to a label entry-wise multi-label version:\begin{equation}
L_{\text{ELR-ML}}(\theta) = \frac{\lambda}{N} \sum_{i=1}^n \sum_{c=1}^C \log \left( 1 - p_{i,c} \cdot y_{i,c} \right).
\end{equation}

This regularization term reinforces early-learned, reliable patterns while attenuating gradients from uncertain or mislabeled entries. Thus, it provides a complementary mechanism to the confidence-based label correction described in the following.

\subsection{Confidence-Based Label Handling}

To address additive and subtractive noise asymmetrically, we propose a confidence-based label correction mechanism that dynamically adapts the supervision applied to each label entry during training. For each class prediction, model confidence (i.e., probability) determines whether a label entry is retained, deactivated, or corrected. Highly confident predictions consistent with the label entry are retained as normal supervision. Predictions indicating moderate uncertainty lead to temporary deactivation, effectively treating the entry as unlabeled in a semi-supervised manner. When the model exhibits high confidence that a label entry is incorrect, the entry is corrected (i.e., flipped) and used as a pseudo-label. This adaptive three-state mechanism enables the model to differentiate between reliable and noisy supervision while maintaining stability in both additive and subtractive noise conditions.

For each label entry \(y_{i,c}\), the correction decision is determined by the predicted confidence \(p_{i,c}\) and four adaptive thresholds: \(t_1^{\text{w0}}\) and \(t_1^{\text{flip}}\) to identify additive noise (false positives) and \(t_0^{\text{w0}}\) and \(t_0^{\text{flip}}\) to identify subtractive noise (false negatives). Let \(\tilde{y}_{i,c}\) denote the corrected label entry and \(w_{i,c}\) its binary loss weight. The correction and weighting rules are:

\[
\tilde{y}_{i,c}, w_{i,c} = 
\begin{cases}
0, 1 & \text{if } y_{i,c} = 1 \text{ and } p_{i,c} < t_1^{\text{flip}}, \\
y_{i,c}, 0 & \text{if } y_{i,c} = 1 \text{ and } t_1^{\text{flip}} \leq p_{i,c} < t_1^{\text{w0}}, \\
1, 1 & \text{if } y_{i,c} = 0 \text{ and } p_{i,c} > t_0^{\text{flip}}, \\
y_{i,c}, 0 & \text{if } y_{i,c} = 0 \text{ and } t_0^{\text{w0}} < p_{i,c} \leq t_0^{\text{flip}}, \\
y_{i,c}, 1 & \text{otherwise}.
\end{cases}
\]

Entries with \(w_{i,c} = 0\) are treated as unlabeled with respect to the confidence-weighted BCE objective and therefore do not contribute to the supervised classification loss during that training iteration. However, when \gls{ELR} is used, the corresponding predictions still contribute to the separate \gls{ELR} regularization term. An illustration of the label handling mechanism is presented in \Cref{fig:method_overview}. The resulting confidence-weighted BCE loss is defined as:

\begin{equation}
\begin{aligned}
L_{\text{BCE-CW}}(\theta) = & \frac{1}{\sum_{i,c} w_{i,c}} \sum_{i=1}^n \sum_{c=1}^C w_{i,c} \cdot \left[ \tilde{y}_{i,c} \log(p_{i,c}) \right. \\
& \left. + (1 - \tilde{y}_{i,c}) \log(1 - p_{i,c}) \right].
\end{aligned}
\end{equation}

The conservative threshold design (i.e., ignoring before flipping) reduces the risk of error accumulation during pseudo-label correction. Label entries are only flipped when the model prediction exceeds a high-confidence threshold, whereas uncertain predictions are temporarily deactivated rather than immediately corrected. This creates a separation between uncertain and highly confident corrections, limiting the propagation of incorrect pseudo-labels during early optimization stages. In addition, the three-state mechanism improves robustness compared to direct correction strategies, since unreliable label entries can be excluded from gradient computation until the model develops more reliable predictions. Together with the regularizing effect of \gls{ELR}, this promotes stable optimization under noisy supervision while reducing over-adaptation to corrupted labels.

The final objective combines confidence-based selective supervision with \gls{ELR}:

\begin{equation}
\begin{aligned}
L_{\text{NAR}}(\theta) = L_{\text{BCE-CW}}(\theta) + L_{\text{ELR-ML}}(\theta).
\end{aligned}
\end{equation}

The first term realizes the semi-supervised component by dynamically weighting and correcting label entries according to confidence, explicitly distinguishing between additive and subtractive noise in the multi-label setting. In contrast, \gls{ELR} acts as a more general regularization term that stabilizes optimization under noisy supervision by discouraging rapid memorization of corrupted labels. Unlike the proposed confidence-based label handling mechanism, \gls{ELR} does not modify the supervisory structure itself, since it neither removes nor corrects label entries. Instead, it complements the proposed three-state mechanism by reducing over-adaptation to unreliable supervision during training.

%% file: sections/4-dataset_design.tex
\begin{figure*}[t!]
    \centering
    \begin{subfigure}{.25\linewidth}
        \dynamiccaption{\caption{}\label{subfigure:example-ucmerced}}{\includegraphics[width=\linewidth]{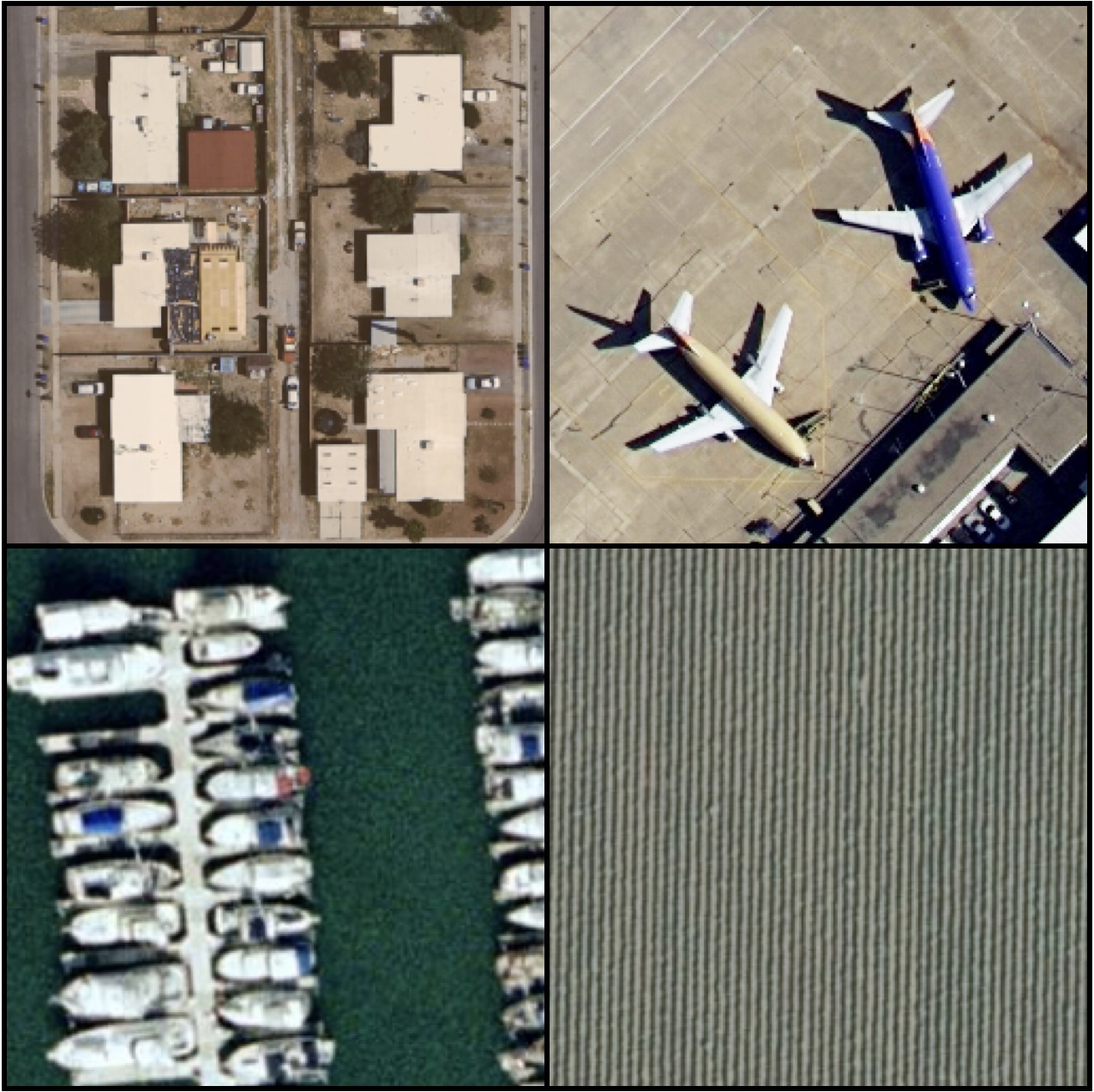}}
    \end{subfigure}\hfil
    \begin{subfigure}{.25\linewidth}
        \dynamiccaption{\caption{}\label{subfigure:example-deepglobe}}{\includegraphics[width=\linewidth]{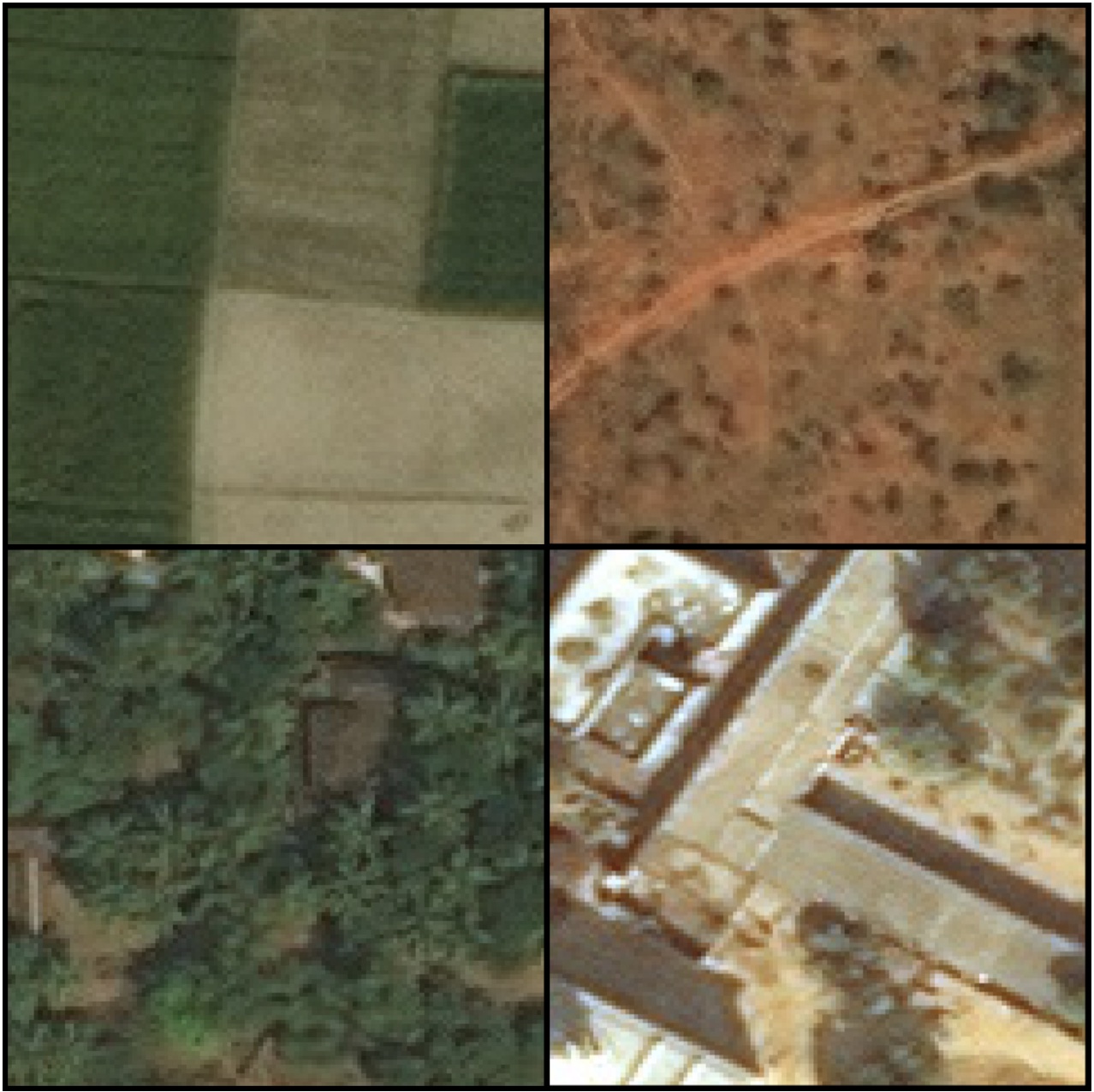}}
    \end{subfigure}\hfil
    \begin{subfigure}{.25\linewidth}
        \dynamiccaption{\caption{}\label{subfigure:example-aidml}}{\includegraphics[width=\linewidth]{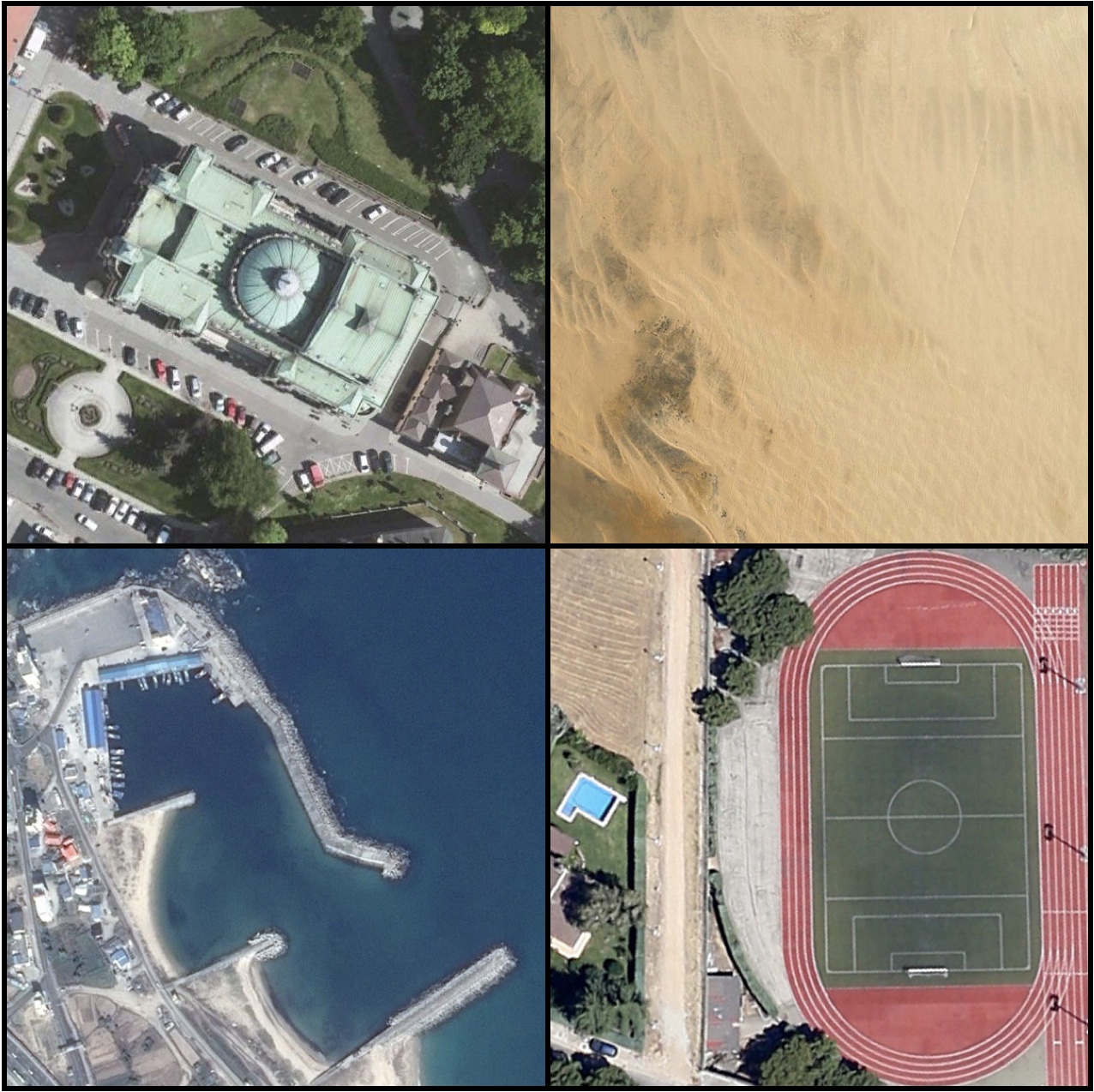}}
    \end{subfigure}
    \caption{Example images of the datasets: \subref{subfigure:example-ucmerced} UCMerced, \subref{subfigure:example-deepglobe} DeepGlobe-ML, and \subref{subfigure:example-aidml} AID-ML.}
    \label{fig:example_images}
\end{figure*}

\begin{table*}[t!]
\caption{Characteristics of the multi-label scene classification datasets used in our experiments.}
\label{table:ds_characteristics}
    \centering
    \renewcommand*{\arraystretch}{1.2}
    \sisetup{group-separator={,}, input-ignore={,}}
     \begin{tabular}{p{3cm} c c c c c c}
     \toprule
    \multirow{2}{*}{Dataset} & {\multirow{2}{*}{$|\mathcal{D}|$}} & {\multirow{2}{*}{\hspace{1.2em}$L$}\hspace{1.2em}} & {Avg. $L$}       &  {\multirow{2}{*}{\hspace{1.2em}$C$\hspace{1.2em}}} & {\multirow{2}{*}{Image Sizes}} & {\multirow{2}{*}{Spatial Resolution}} \\
     & & & {per Image} &  &  &  \\ 
     \midrule
     UCMerced \cite{yang_bag--visual-words_2010} & 1,890 & 17 & 3.3 & 3 & 256$\times$256 & 0.3m \\
     DeepGlobe-ML & 18,185 & 6 & 1.71 & 3 & 120$\times$120 & 0.5m \\ 
     AID-ML \cite{xia_aid_2017} & 2,400 & 30 & 5.15 & 3 & 600$\times$600 & 0.5m-8m\\
     \bottomrule
     \\
    \end{tabular}
\end{table*}

\section{Dataset Description and Experimental Design} \label{dataset_setup}

In the experiments, we have used three RS multi-label datasets: 1) UCMerced Land Use Dataset \cite{chaudhuri_multilabel_2018}, denoted as UCMerced, 2) DeepGlobe-ML, which is a multi-label dataset that we constructed from the DeepGlobe Land Cover Classification Challenge dataset \cite{demir_deepglobe_2018}, and 3) AID-ML \cite{xia_aid_2017}. For each dataset, example images are shown in \cref{fig:example_images}. A comparison of the main characteristics of the considered datasets is provided in \cref{table:ds_characteristics}, while the datasets are briefly described in the following.

\subsection{Datasets}\label{subsec:datasets}

\subsubsection{UCMerced}
The multi-label version of the UCMerced dataset \cite{chaudhuri_multilabel_2018} consists of $2,100$ RGB images that were extracted from the USGS National Map Urban Area Imagery collection for various urban areas around the United States of America. We divided the dataset into $1,890$ train and $210$ test images. The dataset is composed of diverse classes ranging from vehicle classes such as cars, boats, and airplanes to natural land cover classes such as water and trees as well as buildings or pavements. Each image is of size \(256 \times 256\) pixels with a spatial resolution of \SI{0.3}{\metre}. On average, each image is annotated with more than 3 present classes. Each class is present at least 100 times in the dataset. The dataset was manually labeled and guarantees clean annotations. There also exists a single-label version of the dataset \cite{yang_bag--visual-words_2010}. \\

\subsubsection{DeepGlobe-ML}
The original DeepGlobe Land Cover Classification Challenge dataset \cite{demir_deepglobe_2018} is collected from the DigitalGlobe +Vivid Images dataset that contains $1949$ RGB tiles of size $2448\times2448$ pixels with a spatial resolution of \SI{0.5}{\metre} acquired over Thailand, Indonesia, and India. Each tile is associated with a manually annotated ground reference map. The classes comprise urban, agriculture, rangeland, forest, water, barren, and unknown. We constructed the DeepGlobe-ML dataset from the original pixel-level classification dataset. To this end, all tiles were divided into non-overlapping patches of $120\times120$-pixel. The respective multi-labels were then derived from the label information of the associated $120\times120$-pixel reference maps. We discarded all patches containing the class unknown. To construct an appropriate \gls{MLC} dataset, we included all patches containing more than one present class. In addition, we retained only \SI{20}{\percent} of the patches with a single present class, resulting in an average of $1.71$ present classes per patch (cf. $1.28$ without discarding patches). In total, DeepGlobe-ML includes 30443 patches and is split into a training set of 18266 patches, a validation set of 6089 patches, and a test set of 6088 patches. \\

\subsubsection{AID-ML} 
AID-ML \cite{hua_relation_2020} is a multi-label variant derived from the Aerial Image Dataset AID \cite{xia_aid_2017}. The original AID contains $10000$ RGB Google Earth images across $30$ scene categories, each of size $600\times600$ pixels, with spatial resolution ranging from \SI{0.5}{\metre} to \SI{8}{\metre} per pixel. For AID-ML, a subset of $3000$ images was manually re-annotated by visual inspection with object-level labels to support multi-label scene understanding \cite{hua_relation_2020}. The dataset provides $17$ object categories, such as car, building, tree, pavement, bare soil, water, ship, and related object classes. The official split contains $2400$ training images and $600$ test images. The dataset exhibits an average label cardinality of approximately $5.15$ labels per image. \\

\subsection{Experimental Setup}
All experiments are conducted using a ResNet-18 architecture initialized with random weights, implemented via \texttt{PyTorch} \cite{falcon_pytorch_2019}. The models are trained for 30 epochs with the AdamW optimizer and a batch size of 128. We apply a cosine annealing learning rate schedule with an initial learning rate of $1\times10^{-4}$ and 100 warm-up steps. No data augmentation is applied during training, validation, or testing in order to avoid introducing additional confounding factors and to ensure that performance differences can be attributed to the compared noise-robust learning methods. For each configuration, three independent runs are performed, and the reported results represent the mean test set performance in mean average precision (mAP) macro of the best model selected according to validation performance. All experiments are executed on NVIDIA H100 PCIe GPUs with 81 GB of memory.

\subsection{Noise Simulation Strategy for Robustness Evaluation}\label{sec:noise_simulation_strategy}

\begin{figure}[t!]
    \centering
    \includegraphics[width=8.8cm]{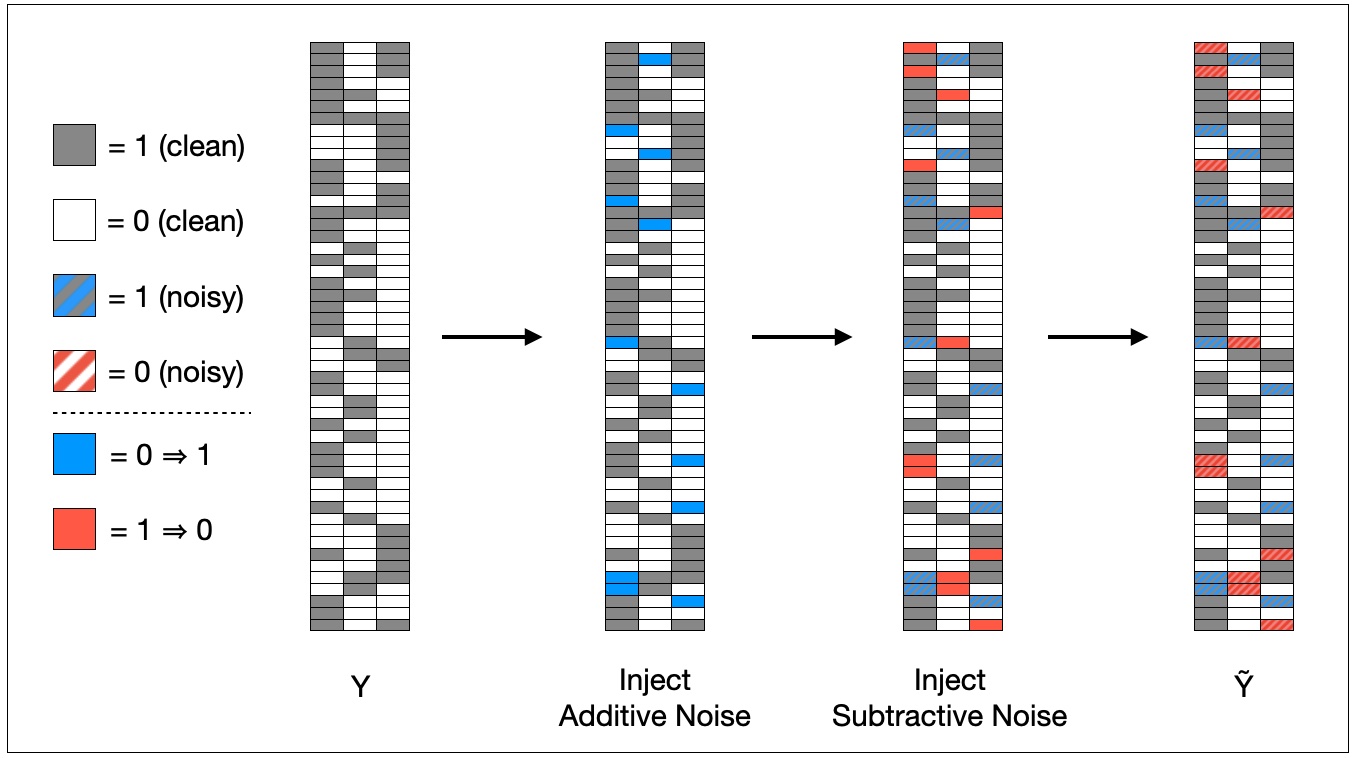}
    \caption{Example of inducing \(20\)\% mixed noise (e.g., \(20\)\% additive noise and 20\% subtractive noise). \(Y\) is the clean label matrix each row representing multi-label \(y_i\), each column representing a class \(c\). \(\tilde{Y}\) represents the noisified label matrix. Additive noise is depicted in blue changing an entry \(c\) in the label \(y_i\) from \(0\) to \(1\), while subtractive noise is depicted in red changing an entry \(c\) in the label \(y_i\) from \(1\) to \(0\). Figure is reproduced from \cite{burgert_effects_2022}.}
    \label{fig:noise_simulation_example}
\end{figure}

Previous works on noise robustness in \gls{MLC} commonly simulate label noise to evaluate robustness by randomly flipping entries in the label matrix without distinguishing between false positives ($0 \rightarrow 1$) and false negatives ($1 \rightarrow 0$) \cite{hua_learning_2020}, \cite{zhao_evaluating_2021}. Since most label entries in multi-label datasets are absent (typically only 10–25\% of the entries are positive, see \Cref{subsec:datasets}), such noise injection results in a strong bias towards false positives, thereby potentially overestimating robustness in operational scenarios under multi-label noise.

\begin{figure*}[t]
    \centering
    \begin{subfigure}{.42\linewidth}
        \dynamiccaption{\caption{}\label{subfigure:oracle_subn}}{\includegraphics[width=\linewidth]{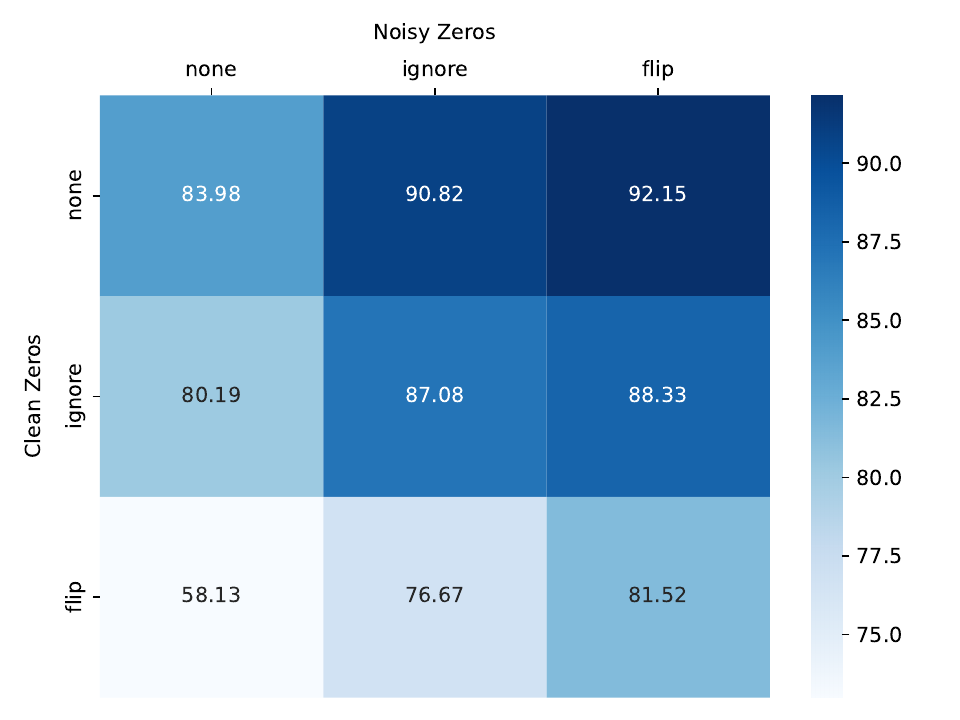}}
    \end{subfigure}
    \begin{subfigure}{.42\linewidth}
        \dynamiccaption{\caption{}\label{subfigure:oracle_addn}}{\includegraphics[width=\linewidth]{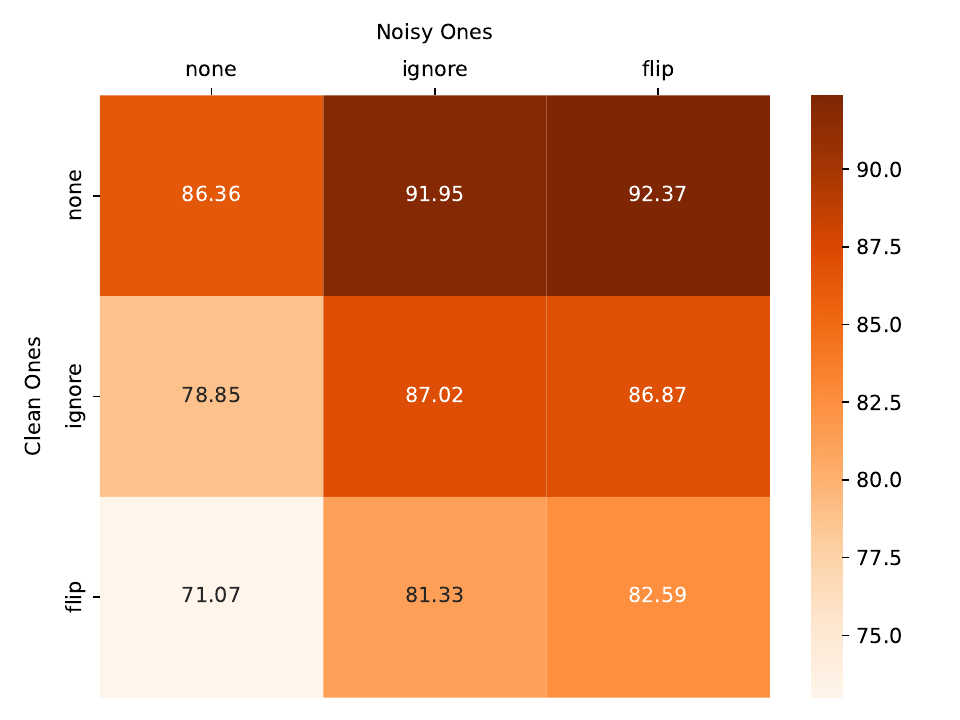}}
    \end{subfigure}
    \caption{Oracle-based evaluation of label handling strategies under 40\% label noise for UCMerced. Each cell shows the mAP macro for combinations of label handling applied to oracle-identified noisy label entries and an equal number of top $k$ most uncertain clean label entries per class. \subref{subfigure:oracle_subn} Subtractive noise. \subref{subfigure:oracle_addn} Additive noise. }
    \label{fig:oracle}
\end{figure*}

To obtain a controlled and disentangled assessment of robustness, we employ the noise simulation strategy introduced by Burgert et al. \cite{burgert_effects_2022}. The strategy distinguishes between three controlled noise conditions: (1) additive noise, which flips a given percentage of negative label entries ($0 \rightarrow 1$) to simulate incorrect present entries; (2) subtractive noise, which flips a given percentage of positive label entries ($1 \rightarrow 0$) to simulate missing present entries; and (3) mixed noise, which applies both in equal proportion, yielding a balanced noise scenario. To ensure comparability across these noise types, the number of injected noisy label entries is normalized by the absolute count of present label entries per class. As a result, classes with more present samples receive proportionally more noisy label entries. The class-wise noise injection preserves the proportions of the class distribution of the training set, preventing artifacts from changed label entry frequencies. An illustration of this procedure is provided in \Cref{fig:noise_simulation_example}. This design enables a separate and balanced evaluation of robustness against false-positive and false-negative perturbations. Hence, it provides a clearer characterization of model behavior under both forms of label noise.

%% file: sections/5-experimental_results.tex
\section{Experimental Results} \label{experimental_results}

The experimental results are organized as follows. In \Cref{sec:label_handling}, we analyze different label handling strategies to examine the impact of label entry deactivation and pseudo-label correction through flipping under controlled noise conditions. This study aims to show that simple pseudo-label flipping is suboptimal in \gls{MLC}, thereby establishing the need for a correction mechanism that adapts to the specific noise type. In \Cref{sec:main_comparison}, we present a comprehensive comparison with existing methods across diverse datasets, noise types, and corruption levels, demonstrating that the proposed method consistently achieves state-of-the-art performance. In \Cref{sec:sensitivity_analysis}, we analyze how the confidence threshold parameters affect the robustness and generalization performance of the proposed method. In \Cref{sec:uniform_noise}, we evaluate the method under the uniform noise assumption to verify its robustness in alternative noise scenarios.  Finally, in \Cref{sec:realworld_noise}, we evaluate the proposed method on a large-scale \gls{RS} dataset with naturally occurring label noise to examine its robustness beyond controlled synthetic corruption settings.

\subsection{Analysis of Label Handling Strategies}\label{sec:label_handling}

To analyze the impact of different label handling strategies and justify the design choices underlying NAR, we conduct an oracle-based experiment on the UCMerced dataset at a noise rate of 40\%. Separate experiments are performed for additive and subtractive noise. For each noise type, an oracle identifies the noisy label entries per class (i.e., 40\% of the present class entries, see \Cref{sec:noise_simulation_strategy} for details). We then evaluate three handling strategies for these identified label entries: no correction (denoted as none), deactivation of their loss contribution (denoted as ignore), and label entry flipping (denoted as flip). To assess the effect of incorrect label noise identification, we repeat this procedure for an equal number of clean label entries per class that are most likely to be mistakenly identified as noisy, based on model uncertainty (i.e., the top $k$ label entries per class with the highest or lowest prediction confidence depending on noise type). This results in a $3 \times 3$ matrix of mAP macro scores per noise type (see \Cref{fig:oracle}).

Consistent with \cite{burgert_effects_2022}, the baseline setting without any correction (i.e., none–none) confirms that subtractive noise generally causes a stronger performance degradation than additive noise. Across both noise types, moving from none to ignore for the oracle-identified noisy label entries substantially improves robustness (e.g., from 83.98 to 90.82 for subtractive and from 86.36 to 91.95 for additive noise), while further switching to flip yields smaller additional gains. This trend indicates that loss deactivation already recovers most of the potential performance lost due to label noise, whereas flipping offers additional benefit only when the noisy label entries are accurately identified and can otherwise lead to performance degradation through unintended correction of clean labels. A similar tendency is observed when examining the uncertain (i.e., likely noisy) clean label entries. Regardless of the specific handling of the noisy label entries, ignoring uncertain clean label entries causes only a moderate decrease in performance, whereas incorrect flipping leads to a much stronger degradation, particularly pronounced for subtractive noise. 

\begin{table*}[t]
\centering
\small
\renewcommand{\arraystretch}{1.2}
\setlength{\tabcolsep}{4pt}
\caption{Performance in mAP macro under different noise types (subtractive, additive, mixed) and noise rates for UCMerced.}
\begin{tabular}{l|ccccc|ccccc|ccccc}
\toprule
\multirow{2}{*}{Method} 
& \multicolumn{5}{c|}{\textbf{Subtractive Noise}} 
& \multicolumn{5}{c|}{\textbf{Additive Noise}} 
& \multicolumn{5}{c}{\textbf{Mixed Noise}} \\
& 10\% & 20\% & 30\% & 40\% & 60\% 
& 10\% & 20\% & 30\% & 40\% & 60\% 
& 10\% & 20\% & 30\% & 40\% & 60\% \\
\midrule
BCE & 89.81 & 88.08 & 85.79 & 83.98 & 77.67 & 90.92 & 89.30 & 87.45 & 86.36 & 81.93 & 87.87 & 80.94 & 74.68 & 66.46 & 50.71 \\
ELR \cite{liu_early-learning_2020} & 89.71 & 89.00 & 87.02 & 85.96 & 81.52 & 91.46 & 90.20 & 89.03 & 87.83 & 84.84 & 89.19 & 84.82 & 78.26 & 69.83 & 53.94 \\
SAT \cite{huang_self-adaptive_2020} & 87.72 & 87.07 & 86.00 & 84.35 & 80.64 & 88.88 & 88.23 & 87.20 & 87.17 & 85.24 & 87.31 & 84.18 & 79.64 & 72.76 & 54.33 \\
ASL \cite{ridnik_asymmetric_2021} & 88.73 & 86.00 & 83.24 & 80.41 & 74.30 & 88.88 & 86.51 & 83.09 & 81.11 & 75.68 & 85.23 & 78.09 & 69.34 & 62.38 & 49.70 \\
BM \cite{song_toward_2024} & 89.89 & 87.80 & 85.55 & 84.20 & 79.55 & 90.41 & 88.87 & 87.61 & 86.44 & 82.51 & 87.74 & 82.73 & 75.88 & 69.06 & 53.42 \\
\midrule
NAR (w/o ELR) & 90.58 & 89.36 & 88.38 & 86.72 & \textbf{83.59} & \textbf{92.67} & \textbf{91.27} & 89.97 & 88.69 & 84.79 & 91.12 & 86.99 & 82.80 & 76.35 & 57.69 \\
NAR (w ELR) & \textbf{91.06} & \textbf{89.67} & \textbf{88.51} & \textbf{87.14} & 83.08 & 92.27 & 91.19 & \textbf{90.30} & \textbf{89.32} & \textbf{85.85} & \textbf{91.77} & \textbf{88.51} & \textbf{83.45} & \textbf{81.15} & \textbf{61.78} \\
\bottomrule
\end{tabular}
\label{tab:results_ucmerced}
\end{table*}

\begin{table*}[t]
\centering
\small
\renewcommand{\arraystretch}{1.2}
\setlength{\tabcolsep}{4pt}
\caption{Performance in mAP macro under different noise types (subtractive, additive, mixed) and noise rates for DeepGlobe-ML.}
\begin{tabular}{l|ccccc|ccccc|ccccc}
\toprule
\multirow{2}{*}{Method} 
& \multicolumn{5}{c|}{\textbf{Subtractive Noise}} 
& \multicolumn{5}{c|}{\textbf{Additive Noise}} 
& \multicolumn{5}{c}{\textbf{Mixed Noise}} \\
& 10\% & 20\% & 30\% & 40\% & 60\%
& 10\% & 20\% & 30\% & 40\% & 60\%
& 10\% & 20\% & 30\% & 40\% & 60\% \\
\midrule
BCE & 76.37 & 73.49 & 72.46 & 71.72 & 67.66 & 77.36 & 75.99 & 75.21 & 74.38 & 73.76 & 75.11 & 73.02 & 63.59 & 59.17 & 51.20 \\
ELR \cite{liu_early-learning_2020} & 78.39 & 76.31 & 75.92 & 74.43 & 71.92 & 78.88 & 76.56 & 74.29 & 74.51 & 72.39 & 76.60 & 71.03 & 65.46 & 59.04 & 48.29 \\
SAT \cite{huang_self-adaptive_2020} & 78.82 & 77.40 & 76.19 & 74.42 & 71.84 & 79.55 & 79.00 & 77.92 & 77.71 & 76.44 & 77.85 & 74.30 & 65.11 & 61.69 & 53.09 \\
ASL \cite{ridnik_asymmetric_2021} & 76.42 & 74.87 & 73.84 & 72.62 & 70.21 & 76.95 & 75.63 & 74.66 & 74.65 & 73.54 & 74.15 & 69.71 & 62.41 & 59.79 & 50.59 \\
BM \cite{song_toward_2024} & 75.15 & 72.48 & 71.65 & 69.26 & 66.78 & 75.51 & 74.04 & 73.75 & 73.32 & 72.62 & 72.53 & 69.44 & 61.65 & 57.37 & 48.62 \\
\midrule
NAR (w/o ELR) & 79.70 & 78.63 & 77.43 & 76.05 & 74.26 & 79.67 & \textbf{79.13} & 78.11 & \textbf{77.88} & \textbf{76.52} & \textbf{79.10} & \textbf{76.33} & \textbf{71.21} & \textbf{65.62} & \textbf{56.64} \\
NAR (w ELR) & \textbf{80.35} & \textbf{79.06} & \textbf{78.36} & \textbf{76.74} & \textbf{75.10} & \textbf{80.15} & 78.84 & \textbf{78.64} & \textbf{77.67} & 76.00 & 79.15 & 75.53 & 70.71 & 64.05 & 56.40 \\
\bottomrule
\end{tabular}
\label{tab:results_deepglobe}
\end{table*}

\begin{table*}[t]
\centering
\small
\renewcommand{\arraystretch}{1.2}
\setlength{\tabcolsep}{4pt}
\caption{Performance in mAP macro under different noise types (subtractive, additive, mixed) and noise rates for AID-ML.}
\begin{tabular}{l|ccccc|ccccc|ccccc}
\toprule
\multirow{2}{*}{Method} 
& \multicolumn{5}{c|}{\textbf{Subtractive Noise}} 
& \multicolumn{5}{c|}{\textbf{Additive Noise}} 
& \multicolumn{5}{c}{\textbf{Mixed Noise}} \\
& 10\% & 20\% & 30\% & 40\% & 60\%
& 10\% & 20\% & 30\% & 40\% & 60\%
& 10\% & 20\% & 30\% & 40\% & 60\% \\
\midrule
BCE & 68.32 & 66.42 & 64.73 & 62.17 & 59.48 & 69.02 & 68.28 & 68.42 & 66.46 & 64.69 & 64.61 & 60.62 & 54.32 & 48.57 & 39.91 \\
ELR \cite{liu_early-learning_2020} & 70.09 & 68.37 & 66.65 & 64.70 & 63.07 & 71.55 & 70.26 & 69.38 & 68.45 & 66.37 & 68.43 & 63.51 & 58.17 & 52.67 & 41.15 \\
SAT \cite{huang_self-adaptive_2020} & 69.49 & 68.15 & 66.24 & 65.49 & 62.86 & 70.29 & 70.17 & 69.97 & 68.46 & 67.27 & 68.78 & 64.62 & 57.26 & 50.53 & 40.58 \\
ASL \cite{ridnik_asymmetric_2021} & 66.21 & 63.40 & 62.14 & 61.47 & 58.18 & 66.20 & 65.46 & 62.65 & 61.55 & 59.49 & 62.73 & 56.43 & 51.65 & 46.88 & 38.69 \\
BM \cite{song_toward_2024} & 68.47 & 67.00 & 65.03 & 63.18 & 61.10 & 69.67 & 69.13 & 68.88 & 68.66 & 66.71 & 67.40 & 62.78 & 57.61 & 51.96 & 40.51 \\
\midrule
NAR (w/o ELR) & 69.91 & 68.54 & 68.49 & 66.86 & 64.27 & 70.61 & 70.16 & 70.10 & 68.52 & 67.53 & 69.13 & 65.53 & 61.77 & 57.56 & 47.10 \\
NAR  (w ELR) & \textbf{70.65} & \textbf{69.49} & \textbf{68.55} & \textbf{67.17} & \textbf{64.84} & \textbf{72.18} & \textbf{71.43} & \textbf{70.76} & \textbf{69.72} & \textbf{67.89} & \textbf{70.49} & \textbf{66.58} & \textbf{64.24} & \textbf{60.75} & \textbf{49.38} \\
\midrule
\end{tabular}
\label{tab:results_aidml}
\end{table*}

In summary, these results demonstrate that a simple pseudo-label flipping rule would lead to severe degradation when clean label entries are inadvertently modified. Approximately 80\% of the recoverable performance can already be achieved through selective deactivation, which provides a safer compromise between robustness and precision. Based on these findings, NAR adopts a three-state mechanism: most noise-suspected label entries are temporarily deactivated, while only those with a high likelihood of being noisy, particularly under subtractive noise, are corrected via pseudo-label flipping.

\subsection{Comparison with Existing Methods}\label{sec:main_comparison}

Building on the label handling analysis, we now present the main comparative evaluation of NAR against existing approaches. The experiments cover the datasets UCMerced, DeepGlobe-ML, and AID-ML under subtractive, additive, and mixed noise with noise rates between 10\% and 60\%. Two variants of the proposed method are considered, NAR without \gls{ELR} and NAR with \gls{ELR}. Baselines include a model trained on a binary cross entropy loss (BCE) without explicit noise regularization, multi-label adaptations of the single-label noise robust methods ELR \cite{liu_early-learning_2020} and SAT \cite{huang_self-adaptive_2020} (see \cite{burgert_effects_2022} for details), the asymmetric loss (ASL) designed for imbalanced multi-label datasets \cite{ridnik_asymmetric_2021}, and BalanceMix (BM) \cite{song_toward_2024}, a mixup-inspired regularization method. A direct comparison with Aksoy et al.\cite{aksoy_multi-label_2022} is omitted, as the design of their method requires the presence of entirely clean samples to function correctly, a condition that is not met in the controlled noise scenarios investigated in this paper. The results are presented per dataset in \Cref{tab:results_ucmerced}, \Cref{tab:results_deepglobe}, and \Cref{tab:results_aidml}.

Under subtractive noise, where present classes are unannotated and thus false negatives are introduced, the NAR variants achieve the highest performance on all three datasets across all evaluated noise rates. On UCMerced, NAR with \gls{ELR} performs best up to a noise rate of 40\%, while NAR without \gls{ELR} achieves the highest result at 60\% (i.e., 83.59). On DeepGlobe-ML and AID-ML, NAR with \gls{ELR} consistently yields the best performance across the entire range of noise rates, with the performance margin increasing as the proportion of noisy label entries grows (e.g., at 60\% noise exceeding ELR by about 3\% and SAT by up to 5\%).

Under additive noise, where absent classes are incorrectly annotated as present and thus false positives are introduced, the relative performance differences between methods are smaller than under subtractive noise. The NAR variants remain consistently competitive, NAR with \gls{ELR} achieving the highest results on UCMerced and AID-ML across all evaluated noise rates. On DeepGlobe-ML, NAR variants generally outperform SAT, though the difference is minimal, often around 0.2\%. These findings are consistent with the oracle analysis in \Cref{sec:label_handling}, where additive noise exhibited the weakest impact on model performance, implying that the potential for further improvement through noise-robust strategies is inherently constrained.

Under mixed noise, representing the simultaneous presence of additive and subtractive noise, performance declines most strongly, particularly beyond a noise rate of 30\%. In this setting, the NAR variants achieve the highest results across all datasets by a difference of up to 8\% compared to the second-best baseline method. On UCMerced and AID-ML, NAR with \gls{ELR} attains the best performance at all noise rates, whereas on DeepGlobe-ML, NAR without \gls{ELR} performs slightly better. This suggests that the primary performance gains stem from the proposed confidence-based three-state supervision mechanism, whereas \gls{ELR} mainly contributes an additional regularization effect whose benefit depends on the dataset characteristics and noise regime. In particular, under very high corruption levels, the proposed supervision adaptation mechanism already performs strong filtering and correction of unreliable label entries, such that the additional regularization introduced by \gls{ELR} does not always lead to further improvements.

Across methods, ELR and SAT consistently improve upon the unregularized BCE baseline, with ELR showing stronger performance under subtractive noise and SAT remaining competitive under additive noise, particularly under higher noise rates. BM exhibits moderate robustness at low corruption levels and under mixed noise for UCMerced and AID-ML, but it does not consistently exceed the BCE baseline, indicating that mixup-based interpolation is not well suited to the spatial and semantic characteristics of \gls{RS} images. The ASL loss yields the lowest performance across all scenarios, even compared to the unregularized BCE baseline, indicating limited robustness to multi-label noise and reduced suitability for the moderate class cardinality of the evaluated \gls{RS} dataset.

In summary, the NAR variants achieve the highest performance in all nine dataset–noise type combinations, consistently outperforming all baselines under the two more severe noise types, subtractive and mixed noise (i.e., improvements between 2\% and 6\%). Only for additive noise on DeepGlobe-ML, the improvements of NAR variants become marginal. These results support the rationale of the proposed approach, showing that the semi-supervised noise-type–adaptive three-state mechanism maintains robustness under various noise conditions.

\subsection{Sensitivity and Parameter Analysis}\label{sec:sensitivity_analysis}

\begin{figure*}[t]
    \centering
    \begin{subfigure}{.32\linewidth}
        \dynamiccaption{\caption{}\label{subfigure:ablation_subn}}{\includegraphics[width=\linewidth]{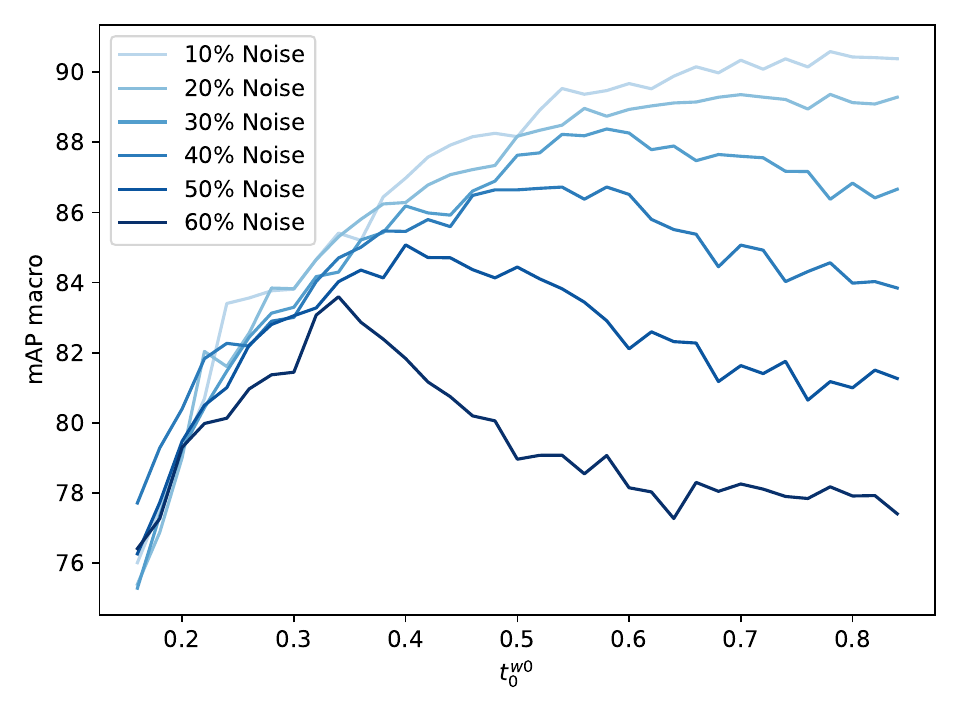}}
    \end{subfigure}
    \begin{subfigure}{.32\linewidth}
        \dynamiccaption{\caption{}\label{subfigure:ablation_addn}}{\includegraphics[width=\linewidth]{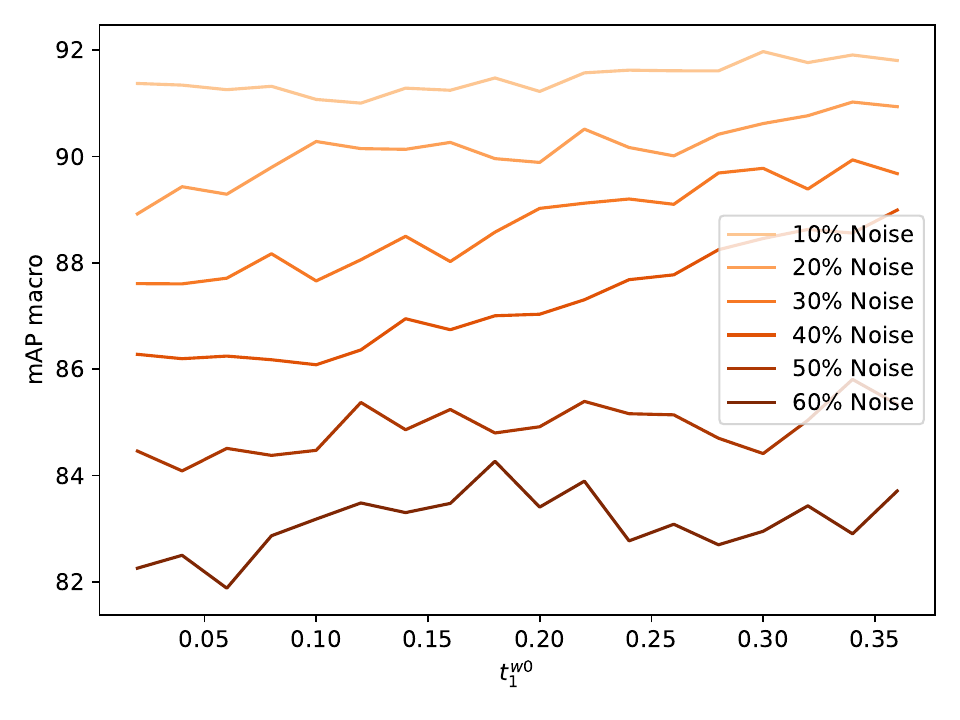}}
    \end{subfigure}
        \begin{subfigure}{.32\linewidth}
        \dynamiccaption{\caption{}\label{subfigure:ablation_mixn}}{\includegraphics[width=\linewidth]{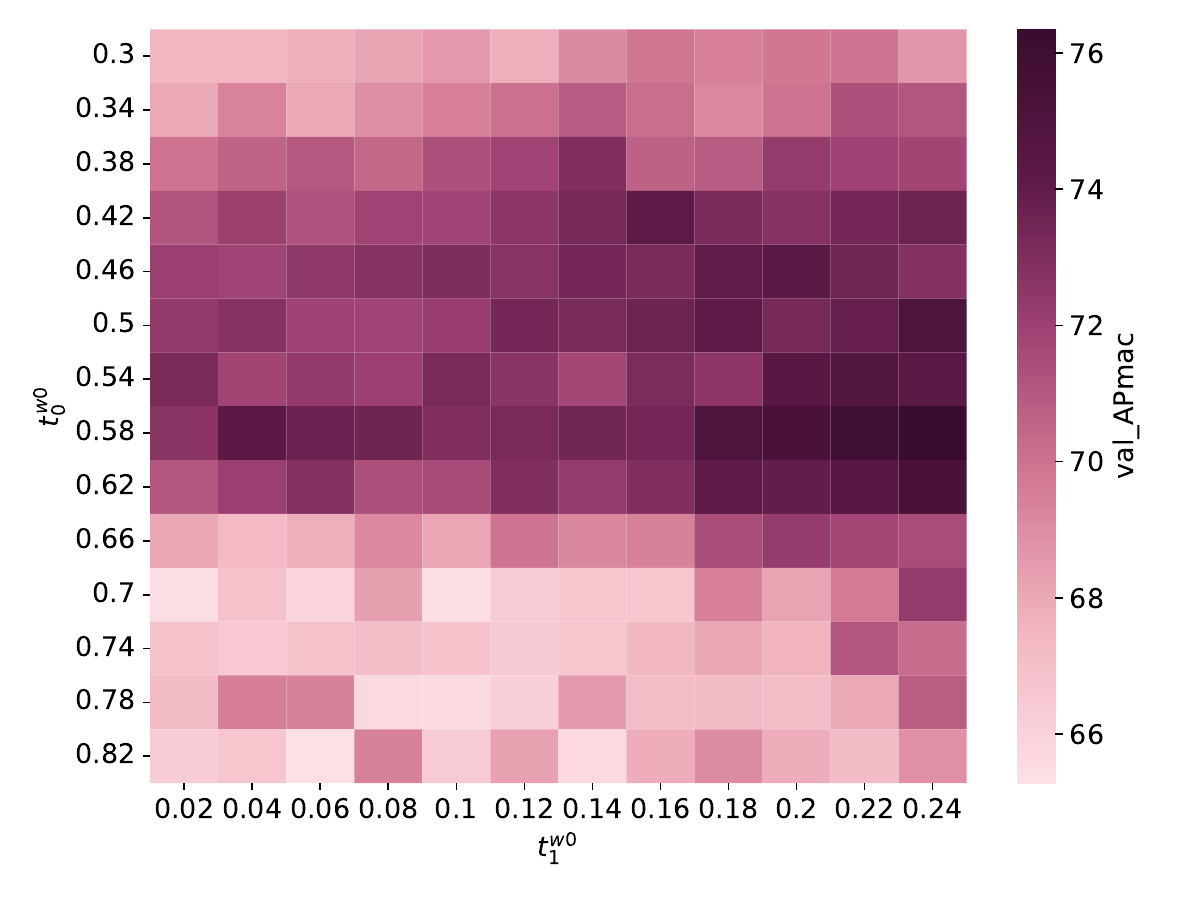}}
    \end{subfigure}
    \caption{Parameter Sensitivity analysis for UCMerced. \subref{subfigure:ablation_subn} Subtractive noise from 10\% to 60\%. \subref{subfigure:ablation_addn} Additive noise from 10\% to 60\%. \subref{subfigure:ablation_mixn} Mixed noise for 40\%.}
    \label{fig:ablation}
\end{figure*}

To examine how the proposed method responds to changes in its main hyperparameters, we analyze the effect of the deactivation thresholds $t_0^{\text{w0}}$ and $t_1^{\text{w0}}$ on the UCMerced dataset under subtractive, additive, and mixed noise. These thresholds determine the confidence intervals that specify when label entries are temporarily excluded from the supervised loss. As shown in \Cref{sec:label_handling}, deactivation constitutes the most decisive factor in achieving robustness, making these parameters central to the behavior of the three-state mechanism.

\Cref{subfigure:ablation_subn}, \Cref{subfigure:ablation_addn}, and \Cref{subfigure:ablation_mixn} present the results across noise types and corruption levels. Under subtractive noise, the optimal $t_0^{\text{w0}}$ decreases with increasing noise rate, indicating that a larger fraction of label entries is correctly identified as unreliable and deactivated at higher corruption levels. At the same time, the performance peak becomes progressively flatter, suggesting that threshold selection has less influence under mild noise but becomes more relevant as label noise intensifies. For additive noise, a similar yet less pronounced pattern is observed, consistent with the generally lower impact of additive noise on model performance explained in \Cref{sec:label_handling}. In the mixed noise scenario, where both error types coexist, $t_1^{\text{w0}}$ exhibits higher sensitivity, and the optimal $t_0^{\text{w0}}$ shifts slightly downward (e.g., from 0.58 under 40\% subtractive noise to 0.50 under 40\% mixed noise).

The sensitivity analysis further indicates that the optimal threshold configuration varies across noise rates. In particular, increasing noise rates generally shift the optimal thresholds toward broader deactivation of uncertain label entries, as a larger fraction of supervision becomes unreliable. More broadly, the optimal threshold configuration also depends on dataset characteristics such as class distributions and label sparsity, which influence the confidence distribution of label entries during training. 

\begin{figure}[t!]
    \centering
    \includegraphics[width=8cm]{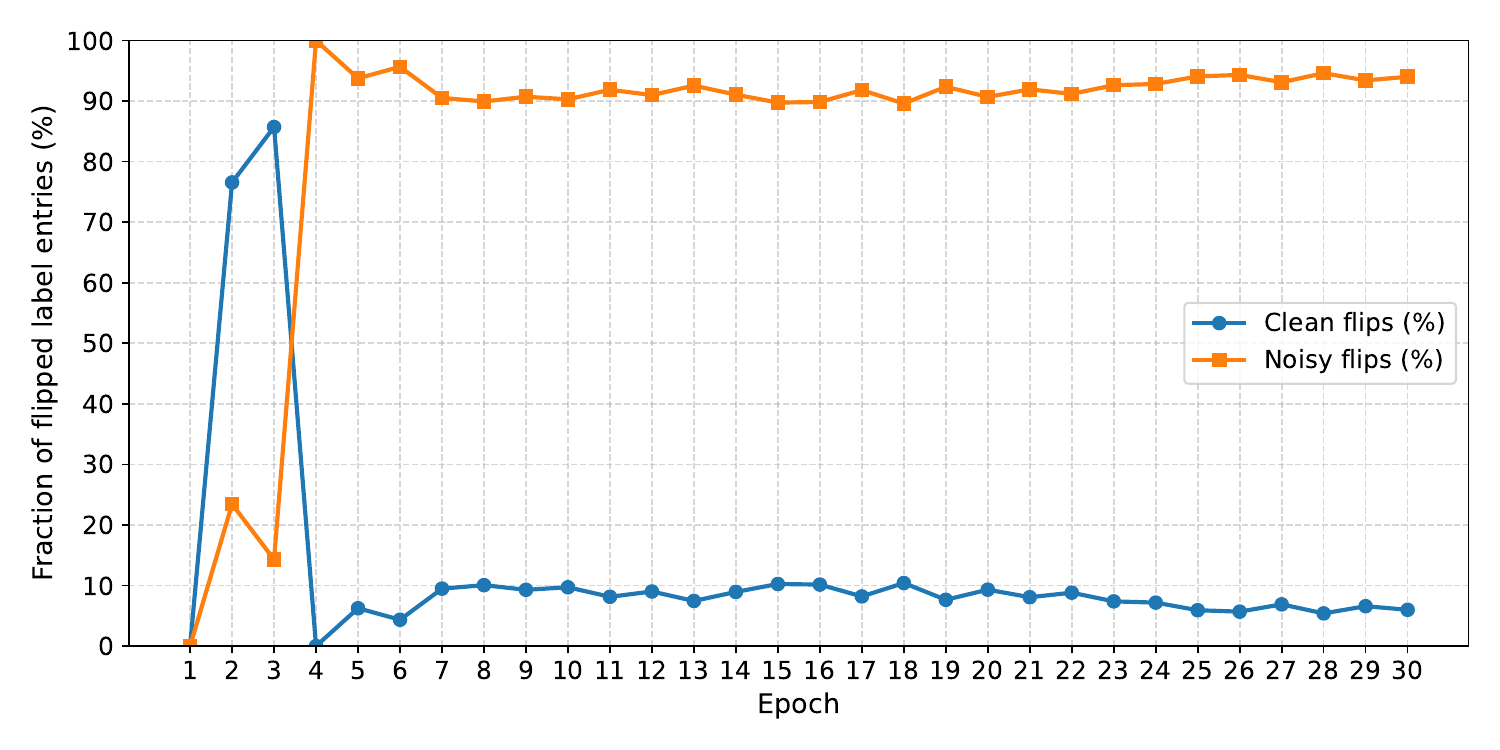}
    \caption{Clean versus noisy flipped label entries across epochs under 40\% subtractive noise for UCMerced.}
    \label{fig:flip_analysis}
\end{figure}

To complement the threshold sensitivity analysis, we additionally analyze the effect of the flipping threshold \(t_0^{\text{flip}}\) by tracking the proportion of flipped clean and noisy label entries during training under 40\% subtractive noise on UCMerced (see \Cref{fig:flip_analysis}). After an initial warm-up and confidence calibration phase during the first epochs, the large majority of flipped label entries correspond to truly noisy labels, whereas the proportion of flips affecting clean label entries remains consistently low (below 10\%). This behavior indicates that the conservative separation between deactivation and flipping effectively limits incorrect label corrections while still allowing the model to recover missing labels.

\subsection{Evaluation Under Uniform Noise Assumption}\label{sec:uniform_noise}

Uniform noise injection has often been used to evaluate robustness of noise robust \gls{MLC} methods \cite{hua_learning_2020}, \cite{zhao_evaluating_2021}, yet it is an inaccurate proxy for operational scenarios as it does not consider label sparsity, class imbalance, and the different effects of additive noise (i.e., false positives) and subtractive noise (i.e., false negatives) discussed in \Cref{sec:noise_simulation_strategy}. Nevertheless, for completeness, we assess performance under this assumption on UCMerced by injecting uniform noise at rates from 10\% to 60\% to evaluate robustness. Given that approximately 20\% of the label matrix entries are positive, uniform noise injection represents a special case of imbalanced mixed noise, with 20\% of the noisy label entries corresponding to subtractive noise and about 80\% of the noisy label entries corresponding to additive noise.

The results in \Cref{tab:results_uniform} demonstrate that the proposed method maintains its robustness under uniform noise. Across all levels of corruption, NAR variants achieve higher performance than the baselines, consistent with the trends observed in the mixed noise experiments in \Cref{sec:main_comparison}. This indicates that the semi-supervised three-state mechanism remains effective even when uniform noise injection inherently produces a predominance of additive noise over subtractive noise.

\begin{table}[h]
\centering
\small
\renewcommand{\arraystretch}{1.2}
\setlength{\tabcolsep}{4pt}
\caption{Performance in mAP macro under uniform noise and different noise rates for UCMerced. Effectively, uniform noise is a special type of mixed noise, 20\% of the noisy label entries correspond to subtractive noise, and 80\% correspond to additive noise.}
\begin{tabular}{l|ccccc}
\toprule
\multirow{2}{*}{Method} 
& \multicolumn{5}{c}{\textbf{Uniform Noise}} \\
& 10\% & 20\% & 30\% & 40\% & 60\% \\
\midrule
BCE & 90.82 & 88.61 & 86.07 & 83.12 & 76.70 \\
ELR \cite{liu_early-learning_2020} & 91.34 & 89.73 & 87.37 & 85.25 & 80.40 \\
SAT \cite{huang_self-adaptive_2020} & 90.00 & 88.45 & 86.81 & 85.60 & 81.48 \\
ASL \cite{ridnik_asymmetric_2021} & 88.51 & 84.97 & 81.87 & 78.08 & 71.76 \\
BM \cite{song_toward_2024} & 90.28 & 87.91 & 85.91 & 83.91 & 78.35 \\
\midrule
NAR (w/o ELR) & 91.70 & 90.12 & 88.46 & 87.42 & 83.19 \\
NAR  (w ELR) & \textbf{92.52} & \textbf{91.15} & \textbf{90.41} & \textbf{88.30} & \textbf{83.36} \\
\midrule
\end{tabular}
\label{tab:results_uniform}
\end{table}

\subsection{Evaluation on Naturally Noisy Data}\label{sec:realworld_noise}

While controlled synthetic noise provides a systematic framework for analyzing robustness under different noise types and noise rates, operational \gls{RS} datasets can naturally contain annotation errors. To evaluate whether NAR generalizes beyond simulated noise, we additionally conduct experiments on BigEarthNet-V2 (BEN-V2) \cite{clasen_reben_2025}, a large-scale \gls{RS} \gls{MLC} dataset known to contain naturally occurring label noise due to its automated annotation process. The results in \Cref{tab:results_benv2} demonstrate that NAR maintains its robustness advantages also under naturally occurring noisy supervision.

Compared to the controlled synthetic experiments, the relative performance differences between methods are smaller on BEN-V2. This is expected, since the effective noise level in BEN-V2 is closer to the lower noise rates considered in the synthetic experiments. Nevertheless, the results suggest that NAR generalizes beyond controlled noise injection settings and remains effective under realistic annotation errors commonly encountered in large-scale \gls{RS} datasets.

\begin{table}[]
\centering
\small
\renewcommand{\arraystretch}{1.2}
\setlength{\tabcolsep}{6pt}
\caption{Performance in mAP macro on the naturally noisy BEN-V2 dataset. Results are averaged across multiple random seeds for the best hyperparameter configuration of each method.}
\begin{tabular}{l|c}
\toprule
{Method} 
& {BENV2} \\
\midrule
BCE & 67.47 \\
ELR \cite{liu_early-learning_2020} & 68.58 \\
SAT \cite{huang_self-adaptive_2020} & 67.47 \\
ASL \cite{ridnik_asymmetric_2021} & 67.83 \\
BM \cite{song_toward_2024} & 66.21 \\
\midrule
NAR (w/o ELR) & 68.34 \\
NAR (w ELR) & \textbf{69.08} \\
\midrule
\end{tabular}
\label{tab:results_benv2}
\end{table}

%% file: sections/6-conclusion.tex
\section{Conclusion}\label{conclusion}

In this paper, we have proposed NAR, a noise-adaptive regularization method for MLC in RS. NAR is designed to explicitly adapt supervision strength at the label entry level according to different multi-label noise types, including additive and subtractive noise, using principles of semi-supervised learning. In particular, it couples confidence-based label handling with \gls{ELR} within a semi-supervised training framework. To this end, NAR employs a three-state mechanism that assesses the reliability of each label entry, retaining entries with high confidence, deactivating those with moderate confidence by treating them as unlabeled, and correcting (i.e., flipping) those with low confidence. An oracle-based analysis has shown that subtractive noise leads to a stronger degradation of model performance than additive noise. It also indicates that selective deactivation of unreliable label entries already recovers most of the attainable performance, whereas extensive label entry flipping can reduce accuracy when clean entries are modified. Experiments on three datasets across three noise types show that NAR consistently achieves the highest performance under subtractive and mixed noise and matches or surpasses competing methods under additive noise, with the performance gap increasing at higher corruption levels. Furthermore, the method also maintains strong performance under the assumption of uniform noise (i.e., imbalanced mixed noise), demonstrating consistent robustness even in unlikely settings of imbalanced mixed noise with dominant additive noise. Overall, these findings demonstrate that adapting the supervision strength to the underlying noise type provides an effective strategy to improve robustness in \gls{RS} \gls{MLC}.

Although the thresholds are manually tuned in the current formulation, calibration strategies based on model confidence statistics or estimated noise levels could further reduce the need for manual parameter tuning. Investigating such calibration strategies therefore represents a promising direction for future work. Another interesting extension of this work is the combination of the proposed semi-supervised, noise-adaptive framework with self-supervised learning approaches in order to improve robustness through stronger feature representations prior to noise-robust training. Finally, future work could investigate controlled synthetic MLC benchmark generation for RS, for example using generative models such as diffusion models, to systematically analyze how class imbalance, present--absent imbalance, and different noise patterns influence the behavior of noise-robust learning methods.